\documentclass[runningheads]{llncs}
\usepackage{booktabs}
\usepackage{cite}
\usepackage{graphicx}
\usepackage{multirow}
\newcommand{\figref}[1]{Fig.\hspace{1mm}\ref{#1}}
\newcommand{\tabref}[1]{Table\hspace{1mm}\ref{#1}}

\begin{document}
\title{Skin lesion classification with ensemble of squeeze-and-excitation networks and semi-supervised learning}
\titlerunning{Skin lesion classification with SENets and semi-supervised learning}
%
\author{Shunsuke KITADA \and Hitoshi IYATOMI}
\authorrunning{S. KITADA and H. IYATOMI}
%
\institute{Major in Applied Informatics, \\ Graduate School of Science and Engineering, \\ Hosei University \\
\email{\{shunsuke.kitada.8y@stu.,iyatomi@\}hosei.ac.jp}}
\maketitle              
\begin{abstract}
 In this report, we introduce the outline of our system in Task 3: Disease Classification of ISIC 2018: Skin Lesion Analysis Towards Melanoma Detection.
 We fine-tuned multiple pre-trained neural network models based on Squeeze-and-Excitation Networks (SENet) which achieved state-of-the-art results in the field of image recognition.
 In addition, we used the mean teachers as a semi-supervised learning framework and introduced some specially designed data augmentation strategies for skin lesion analysis.
 We confirmed our data augmentation strategy improved classification performance and demonstrated 87.2\% in balanced accuracy on the official ISIC2018 validation dataset.
\end{abstract}
\section{Introduction}
The goal of ISIC 2018: Skin Lesion Analysis Towards Melanoma Detection \cite{codella2018skin} is to help participants develop image analysis tools to enable the automated diagnosis of melanoma from dermoscopic images.
The task 3: Disease Classification \cite{tschandl2018ham10000} was to train and predict seven types of skin lesion for each lesion images. The predicted response data is scored by balanced accuracy.
Since the number of cases in each lesion class is biased in the lesion image for training, it is necessary to train a robust classifier that properly classifies under these circumstances.
In this report, we propose skin lesion classification system that fine-tuned multiple Squeeze-and-Excitation Networks \cite{hu2018senet} pre-trained on ImageNet \cite{deng2009imagenet}. 
In this system, we improve the accuracy of model by increasing the variety of data by semi-supervised learning and specially designed data augmentation on imbalance medical data such as skin lesions.

\section{Methodology}

\begin{figure}[htpb]
  \centering
  \includegraphics[scale=0.45]{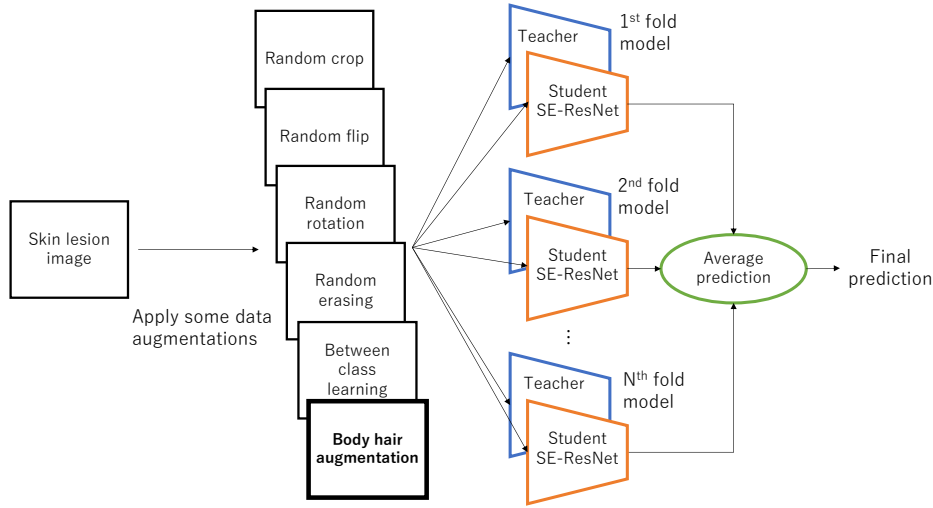}
  \caption{Proposed classification system.}
  \label{fig:proposed_classification_system}
 \end{figure}

\subsection{Squeeze-and-Excitation Networks}
 The outline of our proposed classification system is shown in \figref{fig:proposed_classification_system}.
 The system combines multiple SE-ResNet101 \cite{hu2018senet} models pre-trained on ImageNet \cite{deng2009imagenet}.
 SE-ResNet101 \cite{hu2018senet} is a state-of-the-art model that adopts Squeeze-and-Excitation (SE) block to ResNet101 \cite{he2016deep}, which improves generalization performance by focusing on the channel relationship.
 Matsunaga et al. \cite{matsunaga2017image} has shown that the ensemble of the model is effective also in discrimination of the skin lesion images, we trained $k$ models using $k$-fold cross validation for training data.
 At the time of prediction, the system outputs the result of averaging the predicted values of these models as the final prediction.
 
\subsection{Semi-supervised learning using Mean teacher}
Generally in medical image analysis, the number of malignant data is much smaller than that of benign data, so the final discrimination performance tends to decrease based on imbalanced data.
Therefore, semi-supervised learning framework is introduced in order to learn features from unlabeled data in our classification system.
We introduce mean teacher \cite{tarvainen2017mean} which has achieved state-of-the-art result and learns feature from unlabeled lesion images.

Mean teacher \cite{tarvainen2017mean} uses student model and teacher model.
The student model and teacher model learn based on images given mutually different noises, and the student model learns based on the classification loss and consistency loss.
The classification loss is calculated from the difference between the gold standard label and the prediction by the student.
The consistency loss is difference between predictions by the student and the teacher.
The teacher model is generated by merging the parameters, i.e. network weights, of the student model with the exponential moving average in each epoch.

\subsection{Data augmentation}
In the deep learning methods, it is important to increase diversity of training data.
Therefore, in addition to commonly used data augmentation method \cite{krizhevsky2012imagenet} such as random crop, flip and rotation,
we introduced three distinct data augmentation methods. Two of them showed their superior effect in recent studies,
the other is our new body hair augmentation method that we proved to be effective in discriminating skin lesions.

\subsubsection{Between-class learning}
\begin{figure}[htbp]
 \centering
 \includegraphics[scale=0.5]{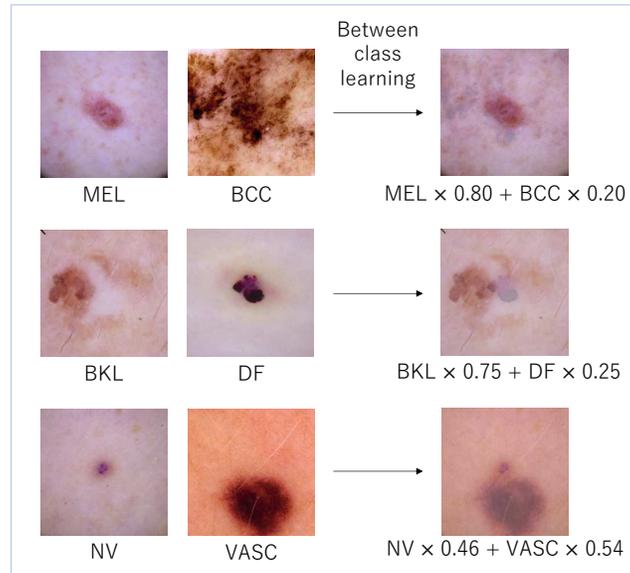}
 \caption{Concept of between-class learning \cite{tokozume2017between}.}
 \label{fig:concept_of_between_class_learning}
\end{figure}

Between-class learning \cite{tokozume2017between} is a data augmentation method that increases the bulk of the overall data across classes by performing linear combination between training data belonging to two different classes.
Concept of between-class learning is shown in \figref{fig:concept_of_between_class_learning}.
In this task, images generated with this method appears more natural than when applied to common image recognition task.

\subsubsection{Random erasing data augmentation}
\begin{figure}[htbp]
 \centering
 \begin{tabular}{c}
  
  \begin{minipage}{0.33\hsize}
   \centering
   \includegraphics[scale=0.35]{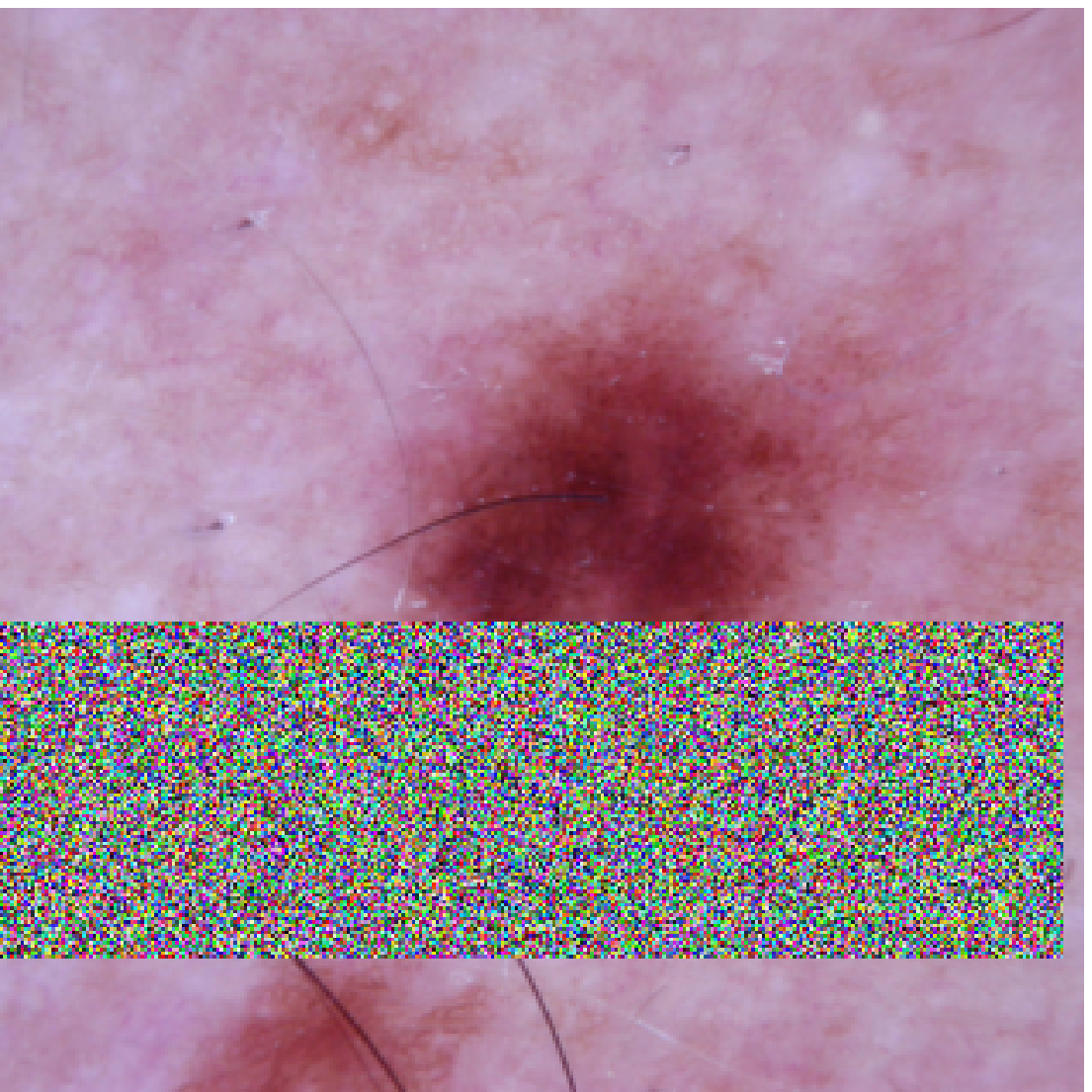}
  \end{minipage}

  \begin{minipage}{0.33\hsize}
   \centering
   \includegraphics[scale=0.35]{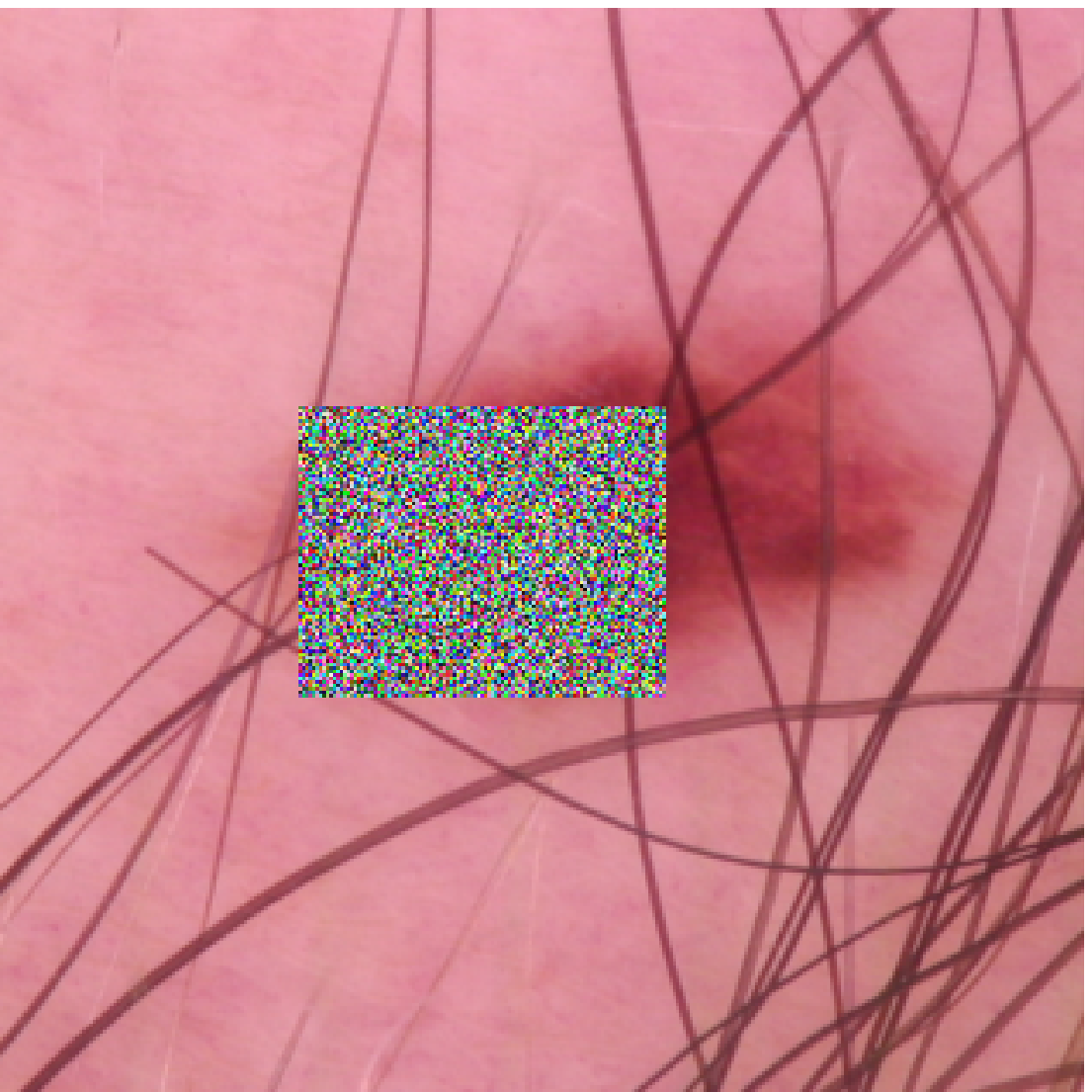}
  \end{minipage}

  \begin{minipage}{0.33\hsize}
   \centering
   \includegraphics[scale=0.35]{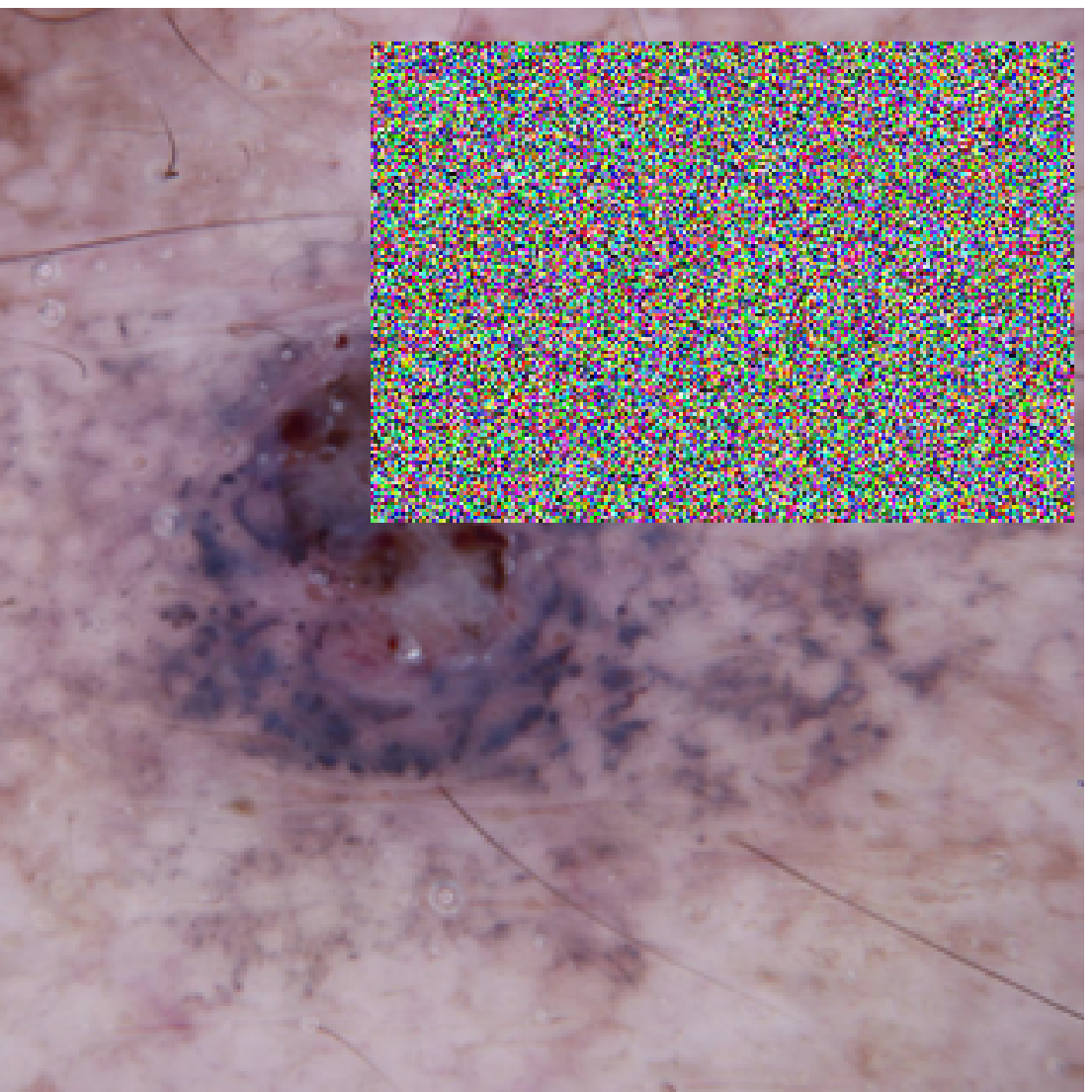}
  \end{minipage}
  
 \end{tabular}
 \caption{Examples of random erasing \cite{zhong2017random}.}
 \label{fig:examples_of_random_erasing}
\end{figure}

Random erasing data augmentation \cite{zhong2017random} is a data augmentation method that randomly selects a rectangle region in an image and erases its pixels with random noise.
Examples of random erasing data augmentation are show in \figref{fig:examples_of_random_erasing}

\subsubsection{Body hair augmentation}
\begin{figure}[htbp]
 \centering
 \includegraphics[scale=0.3]{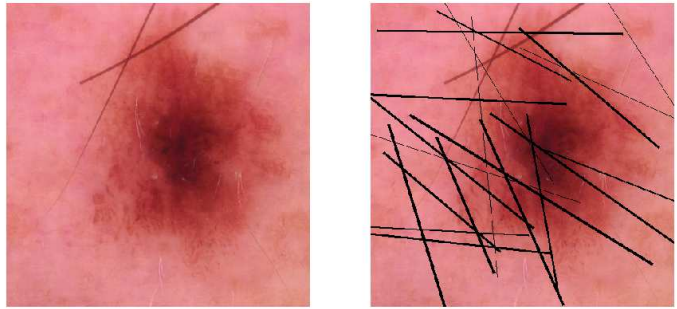}
 \caption{Example of body hair augmentation}
 \label{fig:example_of_body_hair_augmentation}
\end{figure}

Based on our observation, the lesion image shows that there are samples with body hair overlapping in the lesion area.
Several hair removal methodologies have been proposed to address this issue.
However, some problems such as how to interpolate overlapped part were left behind. So we take the opposite approach.
We propose a body hair augmentation which applies pseudo body hair to skin lesion images.
Body hair augmentation is based on Buffon's needele \cite{de1777essai} and gives a line simulating body hair in a pseudo manner.
Example of body hair augmentation is show in \figref{fig:example_of_body_hair_augmentation}.

\section{Experiments and Results}
In the training data of ISIC2018 \cite{codella2018skin, tschandl2018ham10000} shown in \tabref{tab:dataset_of_isic2018}, we trained five SE-ResNet101 model using mean teacher on 5-fold cross validation.
In the training phase, we apply some data augmentations mentioned above randomly (e.g. random crop, flip, rotation, random erasing, body hair augmentation and between-class learning) to the input image, and perform mean subtraction and normalization.
In the prediction phase, we apply test time augmentation \cite{szegedy2015going} and calculate the final prediction result by averaging the predicted values of these five models.
Our classification system is implemented by Chainer \cite{chainer_learningsys2015}.

For our proposed system, we compare balanced accuracy for ISIC 2018 official validation dataset based on the with / without our proposed body hair augmentation in \tabref{tab:result_of_isic2018_validation_dataset}.

\begin{table}[htbp]
 \caption{Dataset of the task 3 of ISIC2018 Challenge}
 \label{tab:dataset_of_isic2018}
 \centering
 \begin{tabular}{ccccccc cc c}
  \toprule
  \multicolumn{7}{c}{labeled}                   & \multicolumn{2}{c}{unlabeled} & \multirow{2}{*}{Total} \\ \cmidrule(lr){1-7} \cmidrule(lr){8-9}
  MEL  & NV   & BCC & AKIEC & BKL  & DF  & VASC & validation       & test       & \\
  \cmidrule(lr){1-7} \cmidrule(lr){8-8} \cmidrule(lr){9-9} \cmidrule(lr){10-10}
  1113 & 6705 & 514 & 327   & 1099 & 115 & 142  & 193              & 1512       & 11720 \\
  \bottomrule
 \end{tabular}
\end{table}

\begin{table}[htbp]
 \caption{Classification performance for the ISIC2018 official validation dataset}
 \label{tab:result_of_isic2018_validation_dataset}
 \centering
 \begin{tabular}[tb]{l c}
  \toprule
  \multicolumn{1}{c}{Model}                            & Balanced accuracy \\ \cmidrule(lr){1-1} \cmidrule(lr){2-2}
  SE-ResNet101 + Mean teacher + body hair augmentation & \textbf{87.2} \\
  SE-ResNet101 + Mean teacher                          & 79.2  \\
  \bottomrule
 \end{tabular}
\end{table}

\section{Conclusions and Future Work}
In this report, we introduce the outline of our method in Task 3: Disease Classification of ISIC 2018: Skin Lesion Analysis Towards Melanoma Detection.
We fine-tuned multiple pre-trained neural network models based on Squeeze-and-Excitation Networks (SENet) which achieved state-of-the-art results in the field of image recognition.
In addition, we used the mean teachers as a semi-supervised learning framework and introduced some specially designed data augmentation strategies for skin lesion analysis.
We confirmed our data augmentation strategy improved classification performance on the official ISIC2018 validation dataset.

In the future, it will be necessary to search for more effective data augmentation for skin lesion images and further study on methods to efficiently handle medical images with unlabeled data.

\section{Acknowledgement}
We thank Daiki SHIMADA for feedback and fruitful discussions and advices.

{\scriptsize
  \bibliographystyle{IEEEtran}
  \bibliography{reference}
}

\end{document}